%% file: KloseMester2019_-_arXiv.tex
\documentclass[sigconf]{acmart}

\usepackage{booktabs} 
\usepackage{amssymb} 
\usepackage{mathtools} 
\usepackage{tabularx} 
\usepackage{subfigure} 
\usepackage{caption} 
\newcolumntype{C}[1]{>{\centering\arraybackslash}p{#1}} 

\setcopyright{rightsretained}

\acmConference[APPIS'19]{Applications of Intelligent Systems}{January 2019}{Las Palmas de Gran Canaria, Spain}
\acmYear{2019}
\copyrightyear{2019}

\begin{document}
\title{Simulated Autonomous Driving in a Realistic Driving Environment using Deep Reinforcement Learning and a Deterministic Finite State Machine}

\author{Patrick Klose}
\affiliation{%
  \department{Visual Sensorics \& Information Processing Lab (CS Dept.)}
  \institution{Goethe University}
  \city{Frankfurt}
 \country{Germany}
}

\author{Rudolf Mester}
\affiliation{%
  \department{Norwegian Open AI Lab \& CS Dept. (IDI)}
 \institution{Norwegian University of Science and Technology}
  \city{Trondheim}
 \country{Norway}
}


\begin{abstract}
In the field of Autonomous Driving, the system controlling the vehicle can be seen as an agent acting in a complex environment and thus naturally fits into the modern framework of Reinforcement Learning.
However, learning to drive can be a challenging task and current results are often restricted to simplified driving environments.
To advance the field, we present a method to adaptively restrict the action space of the agent according to its current driving situation and show that it can be used to swiftly learn to drive in a realistic environment based on the Deep Q-Network algorithm.
\end{abstract}

%
%



\maketitle

\input{body}

\bibliographystyle{ACM-Reference-Format}

\end{document}

%% file: body.tex
\section{Introduction}
In the last decade, the application of Reinforcement Learning (RL) has seen a tremendous growth with producing remarkable results in various fields.
For example, \citeauthor{mnih15} (\citeyear{mnih15}) showed that RL can be used to let an agent learn to play various Atari games beyond the human level, \citeauthor{silver} (\citeyear{silver}) showed that an agent can even learn to beat the world champion in the game of Go and \citeauthor{lillicrap} (\citeyear{lillicrap}) presented that RL can also produce meaningful results when applied to complex control tasks.
This evolution is not only due to the computational resources now available, but also due to the recent theoretical advancement of RL mostly based on combining common RL concepts with deep neural networks (\textit{Deep} RL).

In the field of Autonomous Driving (AD), the RL framework is a promising choice for controlling the vehicle.
In this context, some authors actually showed that RL can be used for learning to drive in yet simplified environments (e.g. \citeauthor{sallab16} \citeyear{sallab16}).
While the results of these studies are remarkable, the demand to make the environment more realistic (e.g. by using a realistic road network) can quickly require an unfeasible amount of training for the agent.

To tackle this problem, we present a novel method, which is a combination of the Deep Q-Network algorithm (DQN; \citeauthor{mnih13} \citeyear{mnih13}) with the concept of a Deterministic Finite State Machine (DFSM; e.g. \citeauthor{hopcroft} \citeyear{hopcroft}), to adaptively restrict the action space of the agent according to its current driving situation. 
While many studies showed that DQN is a powerful algorithm on its own, adaptively restricting the action space decreases the amount of computation needed and incorporates prior knowledge about which actions are \textit{reasonable} in a specific situation to foster exploration and accelerate the training.
We also present the results of an application of our method to show that RL is still an efficient approach to AD, even when the environments are getting more realistic.

\section{Related Work}
Work related to this paper ranges from systems which can only support a human driver to systems which can fully take over the control of a vehicle.
Starting with the former, \citeauthor{desjardins} (\citeyear{desjardins}) proposed an RL system for controlling the (longitudinal) velocity of a vehicle according to another one which is followed.
Similar to this study, \citeauthor{sallab16} (\citeyear{sallab16}) proposed an RL system for lane keeping assistance and showed results of using this system in the racing car simulator TORCS\footnote{https://sourceforge.net/projects/torcs/}.
Furthermore, in another study (\citeauthor{sallab17} \citeyear{sallab17}), the same authors proposed an RL framework for also handling the partial observability of driving situations and the vast amount of information coming from the sensors of a vehicle.

Focusing more on fully autonomous systems, \citeauthor{kuderer} (\citeyear{kuderer}) stated in their study, that there is no common driving style among human drivers and therefore, it is difficult to define \textit{one} global reward function.
To overcome this problem, they showed that inverse RL can be used to previously learn the driving style of a human driver by adjusting parameters of a model which is then used to compute trajectories when in autonomous mode.

Independent from particular driving styles, \citeauthor{shalev-shwartz} (\citeyear{shalev-shwartz}) emphasized that AD is usually a multi-agent problem, which in turn can violate the Markov assumption of RL and that functional safety should be treated separately from it (different methods to tackle these difficulties are presented).
Similar to the previous study, \citeauthor{xiong} (\citeyear{xiong}) combined RL with artificial potential fields and a path tracking method to avoid collisions with other vehicles.

\citeauthor{liaw} (\citeyear{liaw}) used the concept of hierarchical RL and showed that learning a meta-policy composed of already learned policies can be a promising approach to solve problems in the field of AD.
The general idea of using a sub-policy according to the current driving situation of the agent is \textit{conceptually} similar to our approach, since both methods adaptively focus only on a subset of all the possible behavior of the agent.

\section{Theoretical Background}
\subsection{Reinforcement Learning}
In the modern RL framework (\citeauthor{sutton} \citeyear{sutton}) which is often represented as a (finite) Markov Decision Process (MDP), an agent interacts with an environment $\mathcal{E}$ in discrete time steps and at each time step $t$, the agent receives a \textit{representation} of the environment state $S_t \in \mathcal{S}$ (further: state), 
where $\mathcal{S}$ describes the (finite) set of all possible states in $\mathcal{E}$.
Based on $S_t$, the agent chooses from a set of possible actions in that state, i.e., there is a function $\mathcal{A} : \mathcal{S} \to P(\mathcal{A}^+) \setminus \emptyset$ mapping a state to an element of the power set $P$ of $\mathcal{A}^+$ (with the empty set being excluded), whereby $\mathcal{A}^+$ is in turn defined as the (finite) set of all possible actions in $\mathcal{E}$.
Which particular action of $\mathcal{A}(S_t)$ is then chosen is controlled by the (stochastic) policy of the agent defined as the function $\pi(.|S_t) : \mathcal{A}(S_t) \to \mathbb{R}^{\geq0}$, whereby $\sum_{\xi \in \mathcal{A}(S_t)} \pi(\xi|S_t)=1$.
Hence, $\pi$ is a conditional probability distribution, i.e., $\pi(a|S_t)$ represents the probability of taking action $a \in \mathcal{A}(S_t)$ in state $S_t$.

Let $A_t$ be the action, the agent has actually taken in $S_t$.
Then, in time step $t+1$, the agent makes a transition into a new state $S_{t+1}$ and receives a numerical reward $R_{t+1} \in \mathcal{R} \subset \mathbb{R}$ from the (finite) set of all rewards $\mathcal{R}$ as a consequence of its decision about $A_t$ in $S_t$.
These dynamics are described by the (stochastic) environment function $\mathcal{P}(., .|S_t, A_t) : \mathcal{S} \times \mathcal{R} \to \mathbb{R}^{\geq0}$, where it holds that $\sum_{\xi \in \mathcal{S}} \sum_{\zeta \in \mathcal{R}} \mathcal{P}(\xi, \zeta | S_t, A_t) = 1$.
Hence, $\mathcal{P}(s, r|S_t, A_t)$ represents the conditional joint probability for transitioning into state $s$ and receiving reward $r$ when deciding upon action $A_t$ in state $S_t$.
Here, the particular dynamics of the environment in time step $t$ only depend on $S_t$ and $A_t$ (Markov property; \citeauthor{sutton} \citeyear{sutton}).

Let $(R_t, R_{t+1}, ..., R_T)$ be a particular reward sequence from time step $t$ to time step $T$ where it can be assumed that in the case of an episodic task, $T$ is a terminal state with $T<\infty$ and in the case of an infinite task, $T=\infty$ holds (\citeauthor{sutton} \citeyear{sutton}).
The return $G_t$ is then defined as a function over this reward sequence, i.e., $G_t \coloneqq f(R_t, R_{t+1}, ..., R_T)$ and the formal goal of many RL algorithms is to find an optimal policy $\pi^*$ defined as $\pi^* \coloneqq \arg\underset{\pi}{\max}\: \mathbb{E}_{S_{i\geq 0}, R_{j\geq 1} \sim\mathcal{P}, A_{k\geq 0}\sim\pi}(G_1)$, where $S_0$ is drawn from $\mathcal{P}$ which in this case represents a start distribution over $\mathcal{S}$ (\citeauthor{lillicrap} \citeyear{lillicrap}).
One popular definition for $f$, which we also used in this study, is to sum up the discounted future rewards, i.e., $f(R_t, R_{t+1}, ..., R_T) \coloneqq \sum_{\xi=t}^T\gamma^{\xi-t}\cdot R_{\xi}$ with $\gamma \in [0, 1)$ being the discount factor.

To find $\pi^*$, many RL algorithms try to learn the state-action value function $Q_{\pi}(S_t, A_t) \coloneqq \mathbb{E}_{S_{i > t}, R_{j > t} \sim\mathcal{P}, A_{k > t}\sim\pi}(G_{t+1}|S_t, A_t)$ (policy evaluation; \citeauthor{lillicrap} \citeyear{lillicrap}, \citeauthor{sutton} \citeyear{sutton}).
In other words, $Q_{\pi}(S_t, A_t)$ represents the \textit{expected} return of taking action $A_t$ in $S_t$ and subsequently following policy $\pi$ (\citeauthor{sutton} \citeyear{sutton}).
The idea behind this function is that once (approximately) determined for $\pi$, it can be used as guidance to improve the policy towards $\pi^*$ (policy improvement; \citeauthor{sutton} \citeyear{sutton}).

Beside the fact that many RL algorithms have been proven to converge to $\pi^*$ via General Policy Iteration (GPI; \citeauthor{sutton} \citeyear{sutton}), most of them rely on a table-based representation of $Q_{\pi}$ and hence cannot be effectively applied to real world tasks, where at least $\mathcal{S}$ is often very large (\citeauthor{sutton} \citeyear{sutton}).
To overcome this limitation, some RL algorithms insteadly approximate $Q_{\pi}$ based on parameterized functions.
In this context, the DQN algorithm which is also used in this study, is based on ordinary Q-Learning (\citeauthor{watkins} \citeyear{watkins}) but represents $Q_{\pi^*}$ with a neural network.
Furthermore, since working with function approximation in RL can be unstable (\citeauthor{sutton} \citeyear{sutton}), DQN also uses a replay buffer (\citeauthor{lin} \citeyear{lin}) to which the experiences of the agent are saved and from which random mini-batches are drawn for the training.
This method eliminates the strong dependency between subsequent experiences of the agent and shifts the RL problem into the direction of a supervised one for which stable algorithms exist (\citeauthor{mnih13} \citeyear{mnih13}, \citeauthor{mnih15} \citeyear{mnih15}).

\subsection{Deterministic Finite State Machine}
A DFSM is similar to the concept of a MDP, i.e., it is also defined on the basis of states, actions (further: inputs) and related transition dynamics but is fully deterministic and provides no reward signals.
A DFSM is commonly used as a structure to solve various computational problems and can be defined as the 5-tuple $M \coloneqq (W, \Sigma, \delta, w_0, F)$, where $W$ is the (finite) set of states, $\Sigma$ is the (finite) set of inputs, $\delta : W \times \Sigma \to W$ is representing the transition dynamics and $w_0 \in W$ as well as $F \subseteq W$ are describing the starting state and the (finite) set of accepting states, respectively.

Based on this setting, the DFSM starts in $w_0$ and while receiving a sequence of inputs $(\sigma_0, \sigma_1, ..., \sigma_{n-1}) \in \Sigma^n$, it makes transitions according to $\delta$ from state to state.
If the DFSM transitioned after $i$ inputs ($0 < i \leq n$) into an accepting state (e.g. $w_3$ with $w_3 \in F$), then the last $j$ inputs with $0 < j \leq i$ fulfilled the computation and the DFSM is set back to $w_0$.
While using the concept of accepting states is flexible and powerful, we only use the broader concept of a DFSM in this study, i.e., having different states in which some computation can be done as well as having inputs which trigger deterministic transitions between those states.

\begin{table*}[t]
\centering
  \begin{tabular}{C{8cm} C{0.5cm} C{8cm}}
   \includegraphics{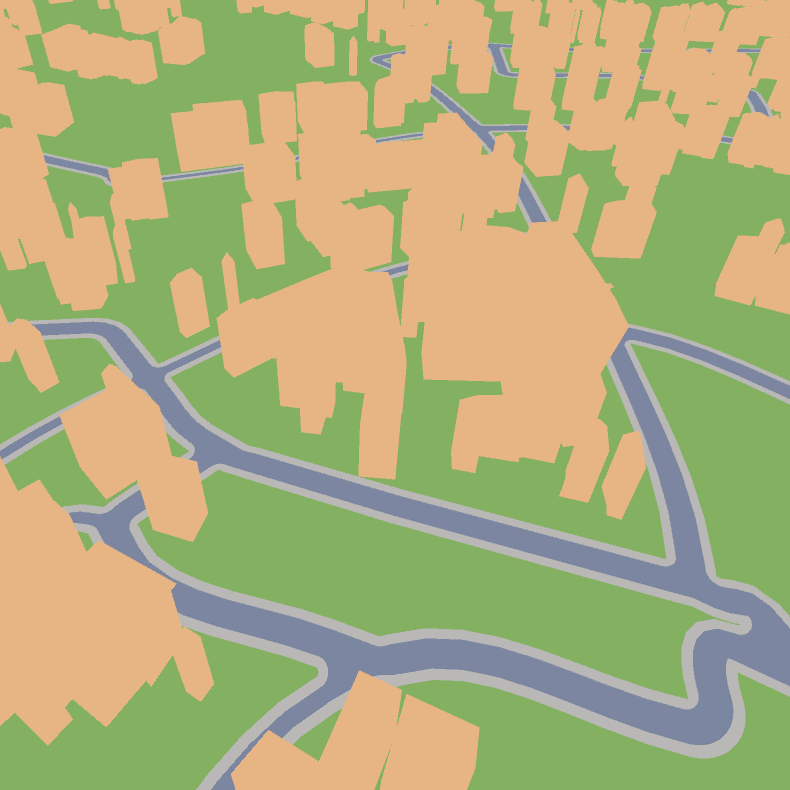}
   \captionof{figure}{An extract from the realistic road network}
 & & 
  \includegraphics{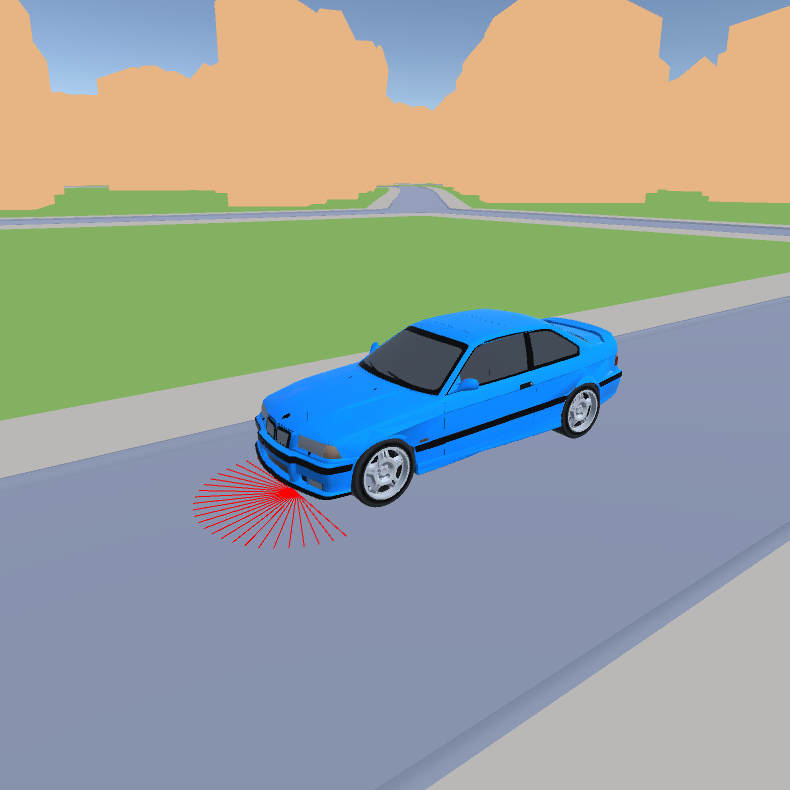}
\captionof{figure}{The vehicle with its simplified LiDAR system}
 \\
\hline
  \end{tabular}
\end{table*}

\section{Driving Environment}
\subsection{Realistic Road Network}
To model the driving environment truly realistic, we decided to use Open Street Maps (OSM)\footnote{https://www.openstreetmap.org} as the data provider for the realistic road network.
Using this data has the advantage that the agent will not only face common and simple but also complex and rare driving situations which may not be present if we would rely on an artificially generated road network.

To map the OSM data into a suitable 3D model, we used the trial version of an OSM to 3D model converter.
Since this study focuses on learning to drive, we decided to map the OSM data directly into a fully semantically-colored 3D model.
This assumes the complementary step of perception to be solved and allows to fully focus on the \emph{acting} problem.

After mapping the OSM data into the 3D model, we used Unity\footnote{https://www.unity.com} as the 3D engine in which the agent has to drive the vehicle.
This software provides a well-engineered physics engine to let the agent also face a physically well-behaving vehicle (see below).

As the region of driving, we chose the small city Gelnhausen\footnote{Geographical coordinates (center): 50.202975N, 9.190527E} which provides all types of roads, i.e., straight and curved roads as well as intersections and roundabouts (an extract is shown in Figure 1).

\subsection{Vehicle}
To implement the vehicle, we decided to use the model of a BMW M3 from Bonecracker Games\footnote{https://www.bonecrackergames.com}.
This model can be seamlessly integrated into Unity and fully uses the built-in physics engine to provide a realistic driving behavior.

The possible inputs to the vehicle are a brake, throttle and steering command with the former two between 0 and 1 and the steering command between -1 and +1.
To simplify the task for the agent, we have implemented a mapping routine merging the brake and throttle command such that a positive value is a throttle command setting the brake command to 0 and a negative value is a brake command setting the throttle command to 0 (the steering command is independent of this routine).

\subsection{Sensors}
To implement the sensors of the vehicle, we decided upon measuring the longitudinal velocity of the vehicle $v\:[\frac{m}{s}]$, its angular velocity $v^{\prime}\:[\frac{rad}{s}]$, and in analogy to a realistic LiDAR system, a number of distances $\vec{d}$ from the vehicle to the curb of the road in $m$ which we call a \textit{circogram} (see Figure 2 for the simplified LiDAR system).
Furthermore, we implemented a collider generating the true-valued boolean collision signal $c$ when the vehicle hit the curb of the road.
This signal is used to let the agent learn to avoid these undesired situations and as a trigger to respawn the vehicle after a collision to one of 64 places distributed across the road network with the selection being randomly (uniform).

\section{Architecture}
\subsection{Overview}
This section provides an overview of the architecture of our system, which we call the Reinforcement Learning based Driving System (RLDS; see Figure 3), to explain the dataflow between the  driving environment and the modules of the agent.
The subsequent sections then provide the implementation details of each module.

While the \textit{Unity} module is providing the driving environment $\mathcal{E}$, all the other modules are belonging to the agent in a broad sense.
In time step $t$, the agent receives from \textit{Unity} the four measurements $v_t$, $v^{\prime}_t$, $\vec{d}_t$, and $c_t$.
The first three values are processed by the \textit{Memorizer} where a history of already received values is maintained.
Using not only immediate measurements but also historical ones can improve the degree to which the Markov property of the states of the agent is fulfilled.
After adding the current measurements to the history, all the currently saved values are transfered to the \textit{State Generator} where the state $S_t$ is generated.

This state is then transfered to the \textit{Navigator} which adaptively generates $\mathcal{A}(S_t)$, i.e., the reasonable actions for the agent in time step $t$.
Both, $S_t$ and $\mathcal{A}(S_t)$ are then forwarded to the \textit{Neural Network} which sequentially calculates (one $a\in \mathcal{A}(S_t)$ and $\hat{Q}_{\pi^*}(S_t, a)$ at a time) the corresponding state-action values.
All these state-action values are then transfered to the \textit{Maximizer} which selects the action $A_t$ with the highest state-action value.
To encourage exploration, $\mathcal{A}(S_t)$ is also forwarded to the \textit{Explorer} which finally decides on the chosen action $A^*_t$ by either staying with $A_t$ or by selecting some random action (uniform) out of $\mathcal{A}(S_t)$.

\begin{table*}[t]
\centering
  \begin{tabular}{C{17.3cm}}
  \includegraphics{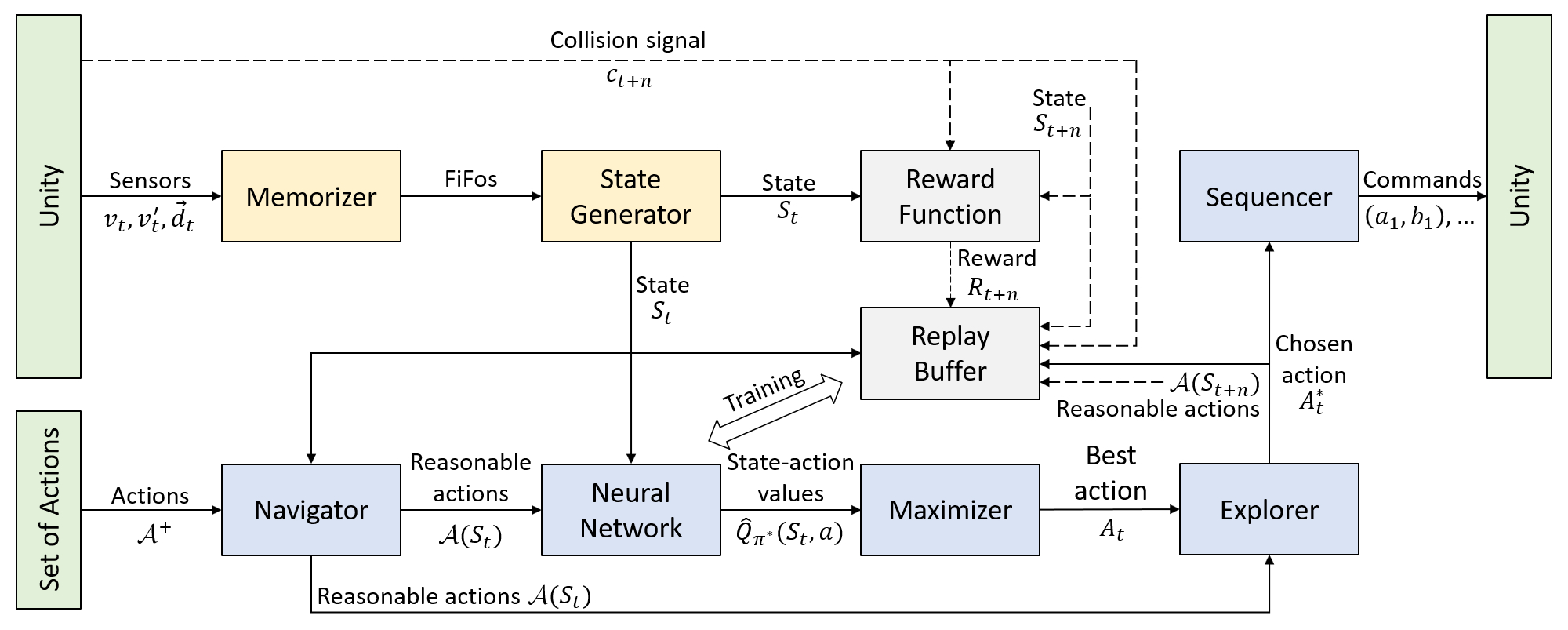}
   \captionof{figure}{The architecture of RLDS}
 \\
\hline
  \end{tabular}
\end{table*}

In either way, the chosen action $A^*_t$ is then sent to the \textit{Sequencer} which creates a sequence of commands ranging from the previous chosen action $A^*_{t-n}$ to the current one (in $n$ time steps) since in the real world, actions can also not be executed without a delay.
Note that after creating the sequence, there will be no further action selection by the agent between time step $t$ and $t+n$ and the agent just sends the commands from the created sequence to \textit{Unity}.

In parallel to this process, the current state $S_t$ and the chosen action $A^*_t$ are also used to build up a SARS tuple\footnote{SARS is an akronym for State, Action, Reward, State.}.
In time step $t+n$, after the \textit{Sequencer} has executed $A^*_t$, it is decided upon $A^*_{t+n}$ based on $S_{t+n}$ and the reward $R_{t+n}$ of the just experienced transition is calculated via the \textit{Reward Function}.
With having $R_{t+n}$ and $S_{t+n}$, the previously started SARS tuple can be completed and saved (together with $c_{t+n}$ and $\mathcal{A}(S_{t+n})$) to the \textit{Replay Buffer}.

Note that since there is no action selection between time step $t$ and $t+n$, the measurements $v_i$, $v^{\prime}_i$, and $\vec{d}_i$ with $t < i < t + n$ are not transfered to the \textit{Memorizer} but simply skipped such that states can only contain measurements saved at time steps where there was an action selection by the agent.
This is important because in any physical system, an action that stimulates the system can only lead to an observable effect if the action is applied for longer than a system-specific time interval.

The actual learning is achieved by adjusting the weights of the \textit{Neural Network} based on the DQN algorithm.

\subsection{Unity}
As already stated, this module is representing the environment $\mathcal{E}$ for the agent and is also defining its corresponding environment function $\mathcal{P}$.
Regarding the vehicle, the simplified LiDAR system has three parameters to adjust, i.e., $\alpha$, $\beta$, and $|\vec{d}|$.
Here, $\alpha \coloneqq 12$ represents the maximum distance of a raycast in $m$, $\beta \coloneqq 180$ represents the angle from the leftmost to the rightmost raycast in $^{\circ}$ and $|\vec{d}| \coloneqq 25$ represents the number of (equally distributed) raycasts.

\subsection{Memorizer and State Generator}
The Memorizer is implemented based on three queues (first-in-first-out) to separately save the historical values of $v$, $v^{\prime}$, and $\vec{d}$.
However, we found that the agent learns best if we set the length of each queue to 1, i.e., when no historical experiences are saved.

The State Generator uses the output of the Memorizer (only the current measurements in this study) and generates the state for the agent.
Here, we decided to use all the measurements available, i.e., $v$, $v^{\prime}$, and $\vec{d}$ as well as a fixed target velocity $\bar{v} \coloneqq 8 [\frac {m} {s}]$ such that the state of the agent in time step $t$ is represented by $[v_t, v^{\prime}_t, \bar{v}, \vec{d}_t]$.

\subsection{Set of Actions}
To exclude extreme commands and to discretize the action space, we decided to provide the agent with 20 values equally distributed between -0.5 and 0.5 for the combined brake and throttle command ($\mathcal{A}^+_1$) and with 100 values equally distributed between -0.8 and 0.8 for the steering command ($\mathcal{A}^+_2$).
Hence, the set of all actions $\mathcal{A}^+ \coloneqq \mathcal{A}^+_1 \times \mathcal{A}^+_2 \subset \mathbb{R}^2$ is represented by 2,000 real-valued pairs.

\subsection{Navigator}
This module is implemented based on the concept of a DFSM and the general idea is to adaptively restrict $\mathcal{A}^+$ by generating $\mathcal{A}(S_t)$ according to $S_t$ and the desired driving direction for the vehicle.

In this context, adaptively restricting the action space has at least three advantages, i.e., first, since our method ensures that $|\mathcal{A}(S_t)| < |\mathcal{A}^+|$ for all the relevant time steps $t$, the corresponding state-action values never have to be calculated for all the 2,000 actions in a time step.
Second, our method allows to use a navigation system guiding the agent along the road network, whereby we kept it as simple as possible in this study, i.e., it decides randomly (uniform) where to drive next when a decision can be made.
Finally, incorporating prior knowledge such that $\mathcal{A}(S_t)$ only contains reasonable actions is meaningful because it accelerates the training (some state-action combinations are not considered at all) and additionally fosters exploration, since the agent is sometimes coerced to handle (approximately) the same state with a different set of actions.

In each time step $t$, in which the agent has to choose an action, the module firstly extracts drivable directions from the circogram $\vec{d}_t$ (contained in $S_t$; see Figure 4) by searching over $\psi \coloneqq 5$ regions for modes consisting of the longest raycasts (eligible modes).
Afterwards, the module decides - based on the DFSM - upon one particular mode and outputs reasonable actions for the agent.
The reason for using a DFSM is that once the module decided upon a particular mode, it should be retained until the agent has driven the vehicle along this part of the road.
Otherwise, when continuously switching between eligible modes, a collision of the vehicle cannot be avoided in many cases.

To implement the DFSM, we decided to use $M \coloneqq (W, \Sigma, \delta, w_0, F)$ with $W \coloneqq \{w_0, w_1, w_2\}$, $\Sigma \coloneqq \{\sigma_0, \sigma_1\}$, 
\begin{equation}
\delta \coloneqq 
\begin{cases}
(w_0, \sigma_0) \to w_0\\
(w_0, \sigma_1) \to w_1\\
(w_1, \sigma_0) \to w_0\\
(w_1, \sigma_1) \to w_2\\
(w_2, \sigma_0) \to w_0\\
(w_2, \sigma_1) \to w_2\\
\end{cases}
\end{equation}
and $F \coloneqq \emptyset$.
Here, $w_0$ represents the state in which the unique eligible mode is selected, $w_1$ represents the state in which one of the multiple eligible modes is randomly (uniform) selected, and $w_2$ represents the state in which the previously chosen mode is selected again.
As the inputs of the DFSM, $\sigma_0$ represents the input signal of having only one eligible mode in the circogram and $\sigma_1$ represents the input signal of having more than one eligible mode.

For instance, let the agent be in time step $t$ while facing a circogram $\vec{d}_t$ containing only one eligible mode as well as let the DFSM be in state $w_0$ (the starting state) such that the DFSM just selects that mode.
If the next circogram $\vec{d}_{t+n}$ contains more than one eligible mode, the DFSM (now in state $w_1$) will randomly (uniform) select one of them.
In subsequent time steps ($t+2n$, $t+3n$, ...), the DFSM will try to reselect this chosen mode as long as the circograms contain more than one eligible mode (state $w_2$) and will otherwise transition back to $w_0$ with behavior as explained.

The reselection of a mode is accomplished by calculating the distances (based on indices) from the center of each eligible mode in the current circogram to the center of the previous chosen mode, whereby the mode with the minimum distance is selected again.
Regarding the generation of reasonable actions for a chosen mode, we decided to only make the steering commands subject to a restriction. 
Hence, the agent can freely choose among the full set of combined brake and throttle commands $\mathcal{A}^+_1$, but only among the restricted set of reasonable steering commands $\mathcal{A}^{+\prime}_2(S_t) \subset \mathcal{A}^+_2$ such that $\mathcal{A}(S_t)$ can be defined as $\mathcal{A}(S_t) \coloneqq \mathcal{A}^+_1 \times \mathcal{A}^{+\prime}_2(S_t)$.

The basis for the definition of $\mathcal{A}^{+\prime}_2(S_t)$ is to additionally separate $\mathcal{A}^+_2$ into the same number $\psi$ of ordered and equally-sized subsets as well as to adaptively define $\mathcal{A}^{+\prime}_2(S_t)$ as the subset of $\mathcal{A}^+_2$ which corresponds to the region of $\vec{d}_t$ containing the center of the currently chosen mode.
In other words, if the desired driving direction (i.e. the chosen mode) is located strongly on the left-hand side in $\vec{d}_t$, then $\mathcal{A}^{+\prime}_2(S_t)$ only contains steering commands for driving left.

\begin{table*}[t]
\centering
  \begin{tabular}{C{8cm} C{0.5cm} C{8cm}}
  \includegraphics{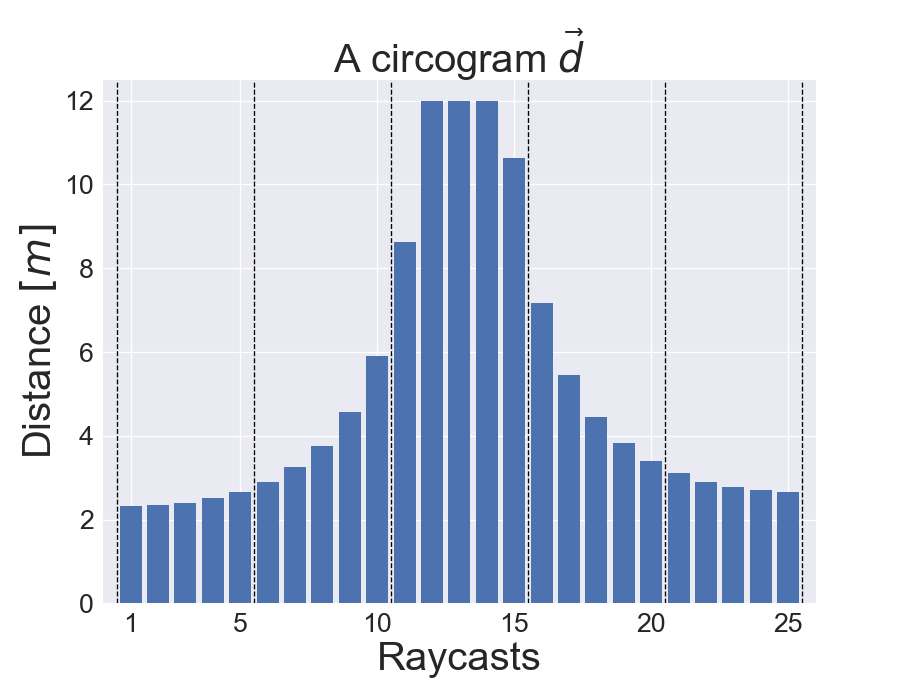}
   \captionof{figure}{A circogram separated into five regions}
 & & 
  \includegraphics{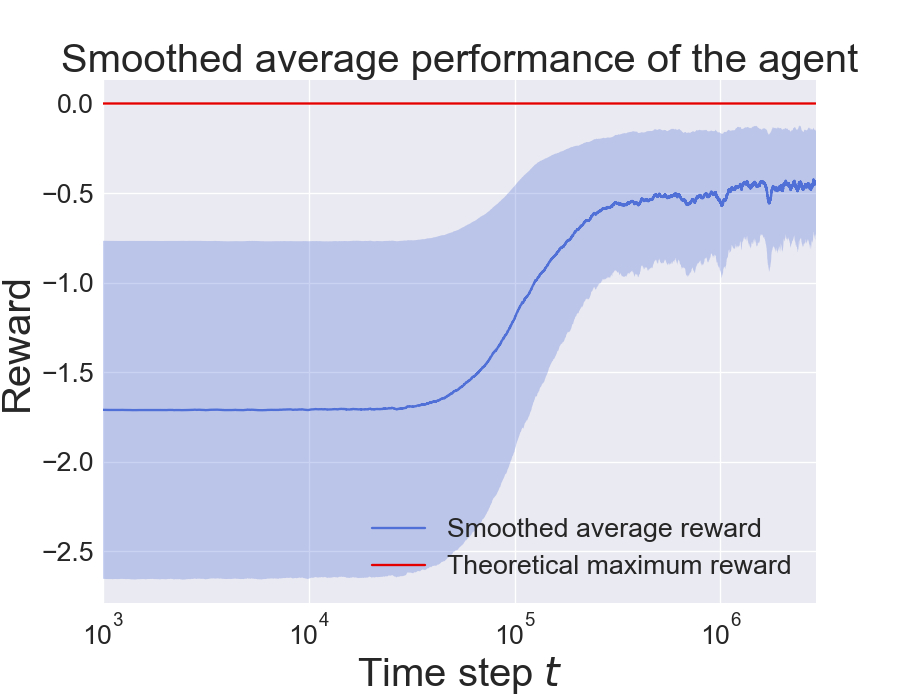}
\captionof{figure}{The smoothed average performance}
 \\
\hline
  \end{tabular}
\end{table*}

\subsection{Neural Network}
To represent the state-action values, we decided upon using a fully-connected neural network with two hidden layers having 400 and 300 neurons, respectively.
The dimension of the input layer was set to 30 and the dimension of the output layer was set to 1.

As the activation functions, we chose LeakyReLU (\citeauthor{maas} \citeyear{maas}) with a negative slope coefficient of 0.3 for the whole neural network except for the output layer, where a linear function was used.
All the weights of the neural network were randomly initialized based on the normal distribution $\mathcal{N}(0, 0.05)$.
For the adjustment of the weights, the mean squared error as the optimization criterion and Adam (\citeauthor{kingma} \citeyear{kingma}) as the optimizer were chosen as well as the learning rate was set to 0.0005.

\subsection{Explorer}
After receiving $A_t$, this module directly returns this action with probability $1 - p_t$ and decides to explore with probability $p_t$.
Between $t=0$ and $t=100,000$, $p_t$ was set to one (full exploration) and it got reduced according to $p_t \coloneqq 0.99999^{(t - 100000)}$ afterwards.

In case of exploring the environment in time step $t$, a random action (uniform) out of $\mathcal{A}^{\prime}(S_t) \subset \mathcal{A}(S_t)$ was selected, whereby $\mathcal{A}^{\prime}(S_t)$ is the subset only containing actions with a combined brake and thottle command larger than zero.
This procedure encourages the agent to explore the behavior of $\mathcal{E}$ when not standing still.

\subsection{Sequencer}
The Sequencer is implemented based on two separate arrays since the action space is two-dimensional.
Let $A^*_{t-n}$ and $A^*_{t}$ be the previously and currently chosen action, let $a_1$ and $b_1$ be the first and second value of the pair $A^*_{t-n}$ as well as let $a_2$ and $b_2$ be the first and second value of the pair $A^*_{t}$.
Then, the Sequencer creates one array leading in $n-1$ steps from $a_1$ to $a_2$, i.e., $[a_1, a_1 + \frac {a_2 - a_1}{n-1}, ..., a_2]$ and another one leading in $n-1$ steps from $b_1$ to $b_2$, i.e., $[b_1, b_1 + \frac {b_2 - b_1}{n-1}, ..., b_2]$ (linear interpolation).

Both arrays are then used for the actual command selection until time step $t+n$ in which two new arrays are generated.
In this study, $n$ was set to 10 which was found to be large enough to provide enough differences between subsequent states and small enough to let the agent control the vehicle with a meaningful frequency.

\subsection{Reward Function}
If the agent let the vehicle collide, the reward function returned a constant value of -20.
Otherwise, the reward for time step $t$ was calculated according to $R_t \coloneqq R^1_t + R^2_t + R^3_t - 3$ with 
\begin{equation}
R^i_t \coloneqq \exp \Big (-0.5 \cdot \Big ( \frac {x^i_t} {\theta_i}  \Big ) ^2 \Big ).
\end{equation}
Here, $x^1_t$ is defined as the absolute difference in the angular velocities between time step $t-n$ and $t$, i.e., $x^1_t \coloneqq |v^{\prime}_{t-n} - v^{\prime}_t|$.
Furthermore, if the width of the road is smaller than $2\cdot\bar{d}$ with $\bar{d}\coloneqq2\;[m]$, $x^2_t$ is defined as the absolute difference between the leftmost and rightmost distance to the curb of the road in time step $t$, i.e., $x^2_t \coloneqq |d^1_t - d^{25}_t|$ and otherwise, $x^2_t$ represents the absolute difference between $\bar{d}$ and the rightmost distance to the curb of the road, i.e., $x^2_t \coloneqq |\bar{d} - d^{25}_t|$.
Finally, $x^3_t$ is defined as the absolute difference between the current and the target velocity of the vehicle in time step $t$, i.e., $x^3_t \coloneqq |v_t - \bar{v}|$.
The parameters $\theta_1$, $\theta_2$, and $\theta_3$ were set to 0.4, 1, and 3, respectively.

\subsection{Remaining Parameters}
We decided to use a warm-up time of 500 SARS tuples to stabilize the training as well as set the batch size for training to 16 SARS tuples and the discount factor $\delta$ to 0.95.
Furthermore, we set the length of the replay buffer to 1,000 SARS tuples.

\section{Results and Conclusion}
Figure 5 shows the smoothed average performance of 10 instances of the agent and it can be seen that its performance improved strongly after starting to decrease the probability of exploration $p$ (due to smoothing, the threshold is slightly shifted to the left).
After about 5 million time steps which corresponds to about 40 hours on a Core i5 with GPU acceleration (GeForce GTX 1070 Ti), the agent controlled the vehicle along the road network nearly collision-free.

These results include learning to (approximately) drive at the target velocity $\bar{v}$ whenever it is possible as well as reducing the velocity of the vehicle before narrow curves and intersections.
Furthermore, we also saw the agent reducing the velocity on straight but narrow roads which is a remarkable result because it requires a proper perception of difficult situations.

However, sometimes the agent let the vehicle still collide, whereby we ascribe this not to the agent as such but rather to the sensors of the agent and the chosen value of $\bar{v}$.
This is because in most of the collisions, $\vec{d}$ does not provided the agent with the necessary information on time to avoid them.

To summarize, we have shown in this study that our method of adaptively restricting the action space of the agent can be used to let an agent swiftly learn to drive a physically well-behaving vehicle nearly collision-free along a realistic road network based on the DQN algorithm.
Worth mentioning is that these results were produced based on an architecture in which almost all of the chosen parameters were not optimized at all.

To conclude, there are at least three avenues for future work, i.e., enhancing the Unity module to provide the agent with additional sensors, completely replacing it by another simulation software, as well as optimizing the parameters to accelerate the training.

\begin{acks}
The authors would like to thank Jacopo Lottero and Hanan Abou Hamdan for developing most of the Unity module.
The authors would also like to thank Christoph Gaudl for implementing the simplified LiDAR system of the vehicle.
\end{acks}

%% file: KloseMester2019_-_arXiv.bbl
\begin{thebibliography}{17}


\ifx \showCODEN    \undefined \def \showCODEN     #1{\unskip}     \fi
\ifx \showDOI      \undefined \def \showDOI       #1{#1}\fi
\ifx \showISBNx    \undefined \def \showISBNx     #1{\unskip}     \fi
\ifx \showISBNxiii \undefined \def \showISBNxiii  #1{\unskip}     \fi
\ifx \showISSN     \undefined \def \showISSN      #1{\unskip}     \fi
\ifx \showLCCN     \undefined \def \showLCCN      #1{\unskip}     \fi
\ifx \shownote     \undefined \def \shownote      #1{#1}          \fi
\ifx \showarticletitle \undefined \def \showarticletitle #1{#1}   \fi
\ifx \showURL      \undefined \def \showURL       {\relax}        \fi
\providecommand\bibfield[2]{#2}
\providecommand\bibinfo[2]{#2}
\providecommand\natexlab[1]{#1}
\providecommand\showeprint[2][]{arXiv:#2}

\bibitem[\protect\citeauthoryear{Desjardins and Chaib-draa}{Desjardins and
  Chaib-draa}{2011}]%
        {desjardins}
\bibfield{author}{\bibinfo{person}{C. Desjardins} {and} \bibinfo{person}{B.
  Chaib-draa}.} \bibinfo{year}{2011}\natexlab{}.
\newblock \showarticletitle{Cooperative Adaptive Cruise Control: A
  Reinforcement Learning Approach}.
\newblock \bibinfo{journal}{\emph{IEEE Transactions on Intelligent
  Transportation Systems}} \bibinfo{volume}{12}, \bibinfo{number}{4}
  (\bibinfo{year}{2011}), \bibinfo{pages}{1248--1260}.
\newblock


\bibitem[\protect\citeauthoryear{Hopcroft, Motwani, and Ullman}{Hopcroft
  et~al\mbox{.}}{2013}]%
        {hopcroft}
\bibfield{author}{\bibinfo{person}{J.~E. Hopcroft}, \bibinfo{person}{R.
  Motwani}, {and} \bibinfo{person}{J.~D. Ullman}.}
  \bibinfo{year}{2013}\natexlab{}.
\newblock \bibinfo{booktitle}{\emph{Introduction to Automata Theory, Languages,
  and Computation}}.
\newblock \bibinfo{publisher}{Pearson Education}.
\newblock


\bibitem[\protect\citeauthoryear{Kingma and Ba}{Kingma and Ba}{2014}]%
        {kingma}
\bibfield{author}{\bibinfo{person}{D.~P. Kingma} {and} \bibinfo{person}{J.~L.
  Ba}.} \bibinfo{year}{2014}\natexlab{}.
\newblock \showarticletitle{Adam: A Method for Stochastic Optimization}.
\newblock \bibinfo{journal}{\emph{arXiv:1412.6980}} (\bibinfo{year}{2014}).
\newblock


\bibitem[\protect\citeauthoryear{Kuderer, Gulati, and Burgard}{Kuderer
  et~al\mbox{.}}{2015}]%
        {kuderer}
\bibfield{author}{\bibinfo{person}{M. Kuderer}, \bibinfo{person}{S. Gulati},
  {and} \bibinfo{person}{W. Burgard}.} \bibinfo{year}{2015}\natexlab{}.
\newblock \showarticletitle{Learning Driving Styles for Autonomous Vehicles
  from Demonstration}.
\newblock \bibinfo{journal}{\emph{Proceedings of the 2015 IEEE International
  Conference on Robotics and Automation (ICRA)}} (\bibinfo{year}{2015}).
\newblock


\bibitem[\protect\citeauthoryear{Liaw, Krishnan, Garg, Crankshaw, Gonzalez, and
  Goldberg}{Liaw et~al\mbox{.}}{2017}]%
        {liaw}
\bibfield{author}{\bibinfo{person}{R. Liaw}, \bibinfo{person}{S. Krishnan},
  \bibinfo{person}{A. Garg}, \bibinfo{person}{D. Crankshaw},
  \bibinfo{person}{J.~E. Gonzalez}, {and} \bibinfo{person}{K. Goldberg}.}
  \bibinfo{year}{2017}\natexlab{}.
\newblock \showarticletitle{Composing Meta-Policies for Autonomous Driving
  Using Hierarchical Deep Reinforcement Learning}.
\newblock \bibinfo{journal}{\emph{arXiv:1711.01503}} (\bibinfo{year}{2017}).
\newblock


\bibitem[\protect\citeauthoryear{Lillicrap, Hunt, Pritzel, Heess, Erez, Tassa,
  Silver, and Wierstra}{Lillicrap et~al\mbox{.}}{2015}]%
        {lillicrap}
\bibfield{author}{\bibinfo{person}{T.~P. Lillicrap}, \bibinfo{person}{J.~J.
  Hunt}, \bibinfo{person}{A. Pritzel}, \bibinfo{person}{N. Heess},
  \bibinfo{person}{T. Erez}, \bibinfo{person}{Y. Tassa}, \bibinfo{person}{D.
  Silver}, {and} \bibinfo{person}{D. Wierstra}.}
  \bibinfo{year}{2015}\natexlab{}.
\newblock \showarticletitle{Continuous Control with Deep Reinforcement
  Learning}.
\newblock \bibinfo{journal}{\emph{1509.02971}} (\bibinfo{year}{2015}).
\newblock


\bibitem[\protect\citeauthoryear{Lin}{Lin}{1993}]%
        {lin}
\bibfield{author}{\bibinfo{person}{L.-J. Lin}.}
  \bibinfo{year}{1993}\natexlab{}.
\newblock \showarticletitle{Reinforcement Learning for Robots Using Neural
  Networks}.
\newblock \bibinfo{journal}{\emph{Dissertation (Carnegie Mellon University
  Pittsburgh)}} (\bibinfo{year}{1993}).
\newblock
\urldef\tempurl%
\url{http://www.dtic.mil/get-tr-doc/pdf?AD=ADA261434}
\showURL{%
\tempurl}


\bibitem[\protect\citeauthoryear{Maas, Hannun, and Ng}{Maas
  et~al\mbox{.}}{2013}]%
        {maas}
\bibfield{author}{\bibinfo{person}{A.~L. Maas}, \bibinfo{person}{A.~Y. Hannun},
  {and} \bibinfo{person}{A.~Y. Ng}.} \bibinfo{year}{2013}\natexlab{}.
\newblock \showarticletitle{Rectifier Nonlinearities Improve Neural Network
  Acoustic Models}.
\newblock \bibinfo{journal}{\emph{Proceedings of the 30th International
  Conference on Machine Learning}} (\bibinfo{year}{2013}).
\newblock


\bibitem[\protect\citeauthoryear{Mnih, Kavukcuoglu, Silver, Graves, Antonoglou,
  Wierstra, and Riedmiller}{Mnih et~al\mbox{.}}{2013}]%
        {mnih13}
\bibfield{author}{\bibinfo{person}{V. Mnih}, \bibinfo{person}{K. Kavukcuoglu},
  \bibinfo{person}{D. Silver}, \bibinfo{person}{A. Graves}, \bibinfo{person}{I.
  Antonoglou}, \bibinfo{person}{D. Wierstra}, {and} \bibinfo{person}{M.
  Riedmiller}.} \bibinfo{year}{2013}\natexlab{}.
\newblock \showarticletitle{Playing Atari with Deep Reinforcement Learning}.
\newblock \bibinfo{journal}{\emph{arXiv:1312.5602}} (\bibinfo{year}{2013}).
\newblock


\bibitem[\protect\citeauthoryear{Mnih, Kavukcuoglu, Silver, Rusu, Veness,
  Bellemare, Graves, Riedmiller, Fidjeland, Ostrovski, Petersen, Beattie,
  Sadik, Antonoglou, King, Kumaran, Wierstra, Legg, and Hassabis}{Mnih
  et~al\mbox{.}}{2015}]%
        {mnih15}
\bibfield{author}{\bibinfo{person}{V. Mnih}, \bibinfo{person}{K. Kavukcuoglu},
  \bibinfo{person}{D. Silver}, \bibinfo{person}{A.~A. Rusu},
  \bibinfo{person}{J. Veness}, \bibinfo{person}{M.~G. Bellemare},
  \bibinfo{person}{A. Graves}, \bibinfo{person}{M. Riedmiller},
  \bibinfo{person}{A.~K. Fidjeland}, \bibinfo{person}{G. Ostrovski},
  \bibinfo{person}{S. Petersen}, \bibinfo{person}{C. Beattie},
  \bibinfo{person}{A. Sadik}, \bibinfo{person}{I. Antonoglou},
  \bibinfo{person}{H. King}, \bibinfo{person}{D. Kumaran}, \bibinfo{person}{D.
  Wierstra}, \bibinfo{person}{S. Legg}, {and} \bibinfo{person}{D. Hassabis}.}
  \bibinfo{year}{2015}\natexlab{}.
\newblock \showarticletitle{Human-level Control through Deep Reinforcement
  Learning}.
\newblock \bibinfo{journal}{\emph{Nature}} \bibinfo{volume}{518},
  \bibinfo{number}{7540} (\bibinfo{year}{2015}), \bibinfo{pages}{529–533}.
\newblock


\bibitem[\protect\citeauthoryear{Sallab, Abdou, Perot, and Yogamani}{Sallab
  et~al\mbox{.}}{2016}]%
        {sallab16}
\bibfield{author}{\bibinfo{person}{A.~E. Sallab}, \bibinfo{person}{M. Abdou},
  \bibinfo{person}{E. Perot}, {and} \bibinfo{person}{S. Yogamani}.}
  \bibinfo{year}{2016}\natexlab{}.
\newblock \showarticletitle{End-to-End Deep Reinforcement Learning for Lane
  Keeping Assist}.
\newblock \bibinfo{journal}{\emph{arXiv:1612.04340}} (\bibinfo{year}{2016}).
\newblock


\bibitem[\protect\citeauthoryear{Sallab, Abdou, Perot, and Yogamani}{Sallab
  et~al\mbox{.}}{2017}]%
        {sallab17}
\bibfield{author}{\bibinfo{person}{A.~E. Sallab}, \bibinfo{person}{M. Abdou},
  \bibinfo{person}{E. Perot}, {and} \bibinfo{person}{S. Yogamani}.}
  \bibinfo{year}{2017}\natexlab{}.
\newblock \showarticletitle{Deep Reinforcement Learning Framework for
  Autonomous Driving}.
\newblock \bibinfo{journal}{\emph{Electronic Imaging, Autonomous Vehicles and
  Machines}}  \bibinfo{volume}{7} (\bibinfo{year}{2017}),
  \bibinfo{pages}{70--76}.
\newblock


\bibitem[\protect\citeauthoryear{Shalev-Shwartz, Shammah, and
  Shashua}{Shalev-Shwartz et~al\mbox{.}}{2016}]%
        {shalev-shwartz}
\bibfield{author}{\bibinfo{person}{S. Shalev-Shwartz}, \bibinfo{person}{S.
  Shammah}, {and} \bibinfo{person}{A. Shashua}.}
  \bibinfo{year}{2016}\natexlab{}.
\newblock \showarticletitle{Safe, Multi-Agent, Reinforcement Learning for
  Autonomous Driving}.
\newblock \bibinfo{journal}{\emph{arXiv:1610.03295}} (\bibinfo{year}{2016}).
\newblock


\bibitem[\protect\citeauthoryear{Silver, Huang, Maddison, Guez, Sifre, van~den
  Driessche, Schrittwieser, Antonoglou, Panneershelvam, Lanctot, Dieleman,
  Grewe, Nham, Kalchbrenner, Sutskever, Lillicrap, Leach, Kavukcuoglu, Graepel,
  and Hassabis}{Silver et~al\mbox{.}}{2016}]%
        {silver}
\bibfield{author}{\bibinfo{person}{D. Silver}, \bibinfo{person}{A. Huang},
  \bibinfo{person}{C.~J. Maddison}, \bibinfo{person}{A. Guez},
  \bibinfo{person}{L. Sifre}, \bibinfo{person}{G. van~den Driessche},
  \bibinfo{person}{J. Schrittwieser}, \bibinfo{person}{I. Antonoglou},
  \bibinfo{person}{V. Panneershelvam}, \bibinfo{person}{M. Lanctot},
  \bibinfo{person}{S. Dieleman}, \bibinfo{person}{D. Grewe},
  \bibinfo{person}{J. Nham}, \bibinfo{person}{N. Kalchbrenner},
  \bibinfo{person}{I. Sutskever}, \bibinfo{person}{T. Lillicrap},
  \bibinfo{person}{M. Leach}, \bibinfo{person}{K. Kavukcuoglu},
  \bibinfo{person}{T. Graepel}, {and} \bibinfo{person}{D. Hassabis}.}
  \bibinfo{year}{2016}\natexlab{}.
\newblock \showarticletitle{Mastering the Game of Go with Deep Neural Networks
  and Tree Search}.
\newblock \bibinfo{journal}{\emph{Nature}} \bibinfo{volume}{529},
  \bibinfo{number}{7587} (\bibinfo{year}{2016}), \bibinfo{pages}{484--489}.
\newblock


\bibitem[\protect\citeauthoryear{Sutton and Barto}{Sutton and Barto}{1998}]%
        {sutton}
\bibfield{author}{\bibinfo{person}{R.~S. Sutton} {and} \bibinfo{person}{A.~G.
  Barto}.} \bibinfo{year}{1998}\natexlab{}.
\newblock \bibinfo{booktitle}{\emph{Reinforcement Learning: An Introduction}}.
\newblock \bibinfo{publisher}{The MIT Press}.
\newblock


\bibitem[\protect\citeauthoryear{Watkins}{Watkins}{1989}]%
        {watkins}
\bibfield{author}{\bibinfo{person}{C.~J. C.~H. Watkins}.}
  \bibinfo{year}{1989}\natexlab{}.
\newblock \showarticletitle{Learning from Delayed Rewards}.
\newblock \bibinfo{journal}{\emph{Dissertation (King's College London)}}
  (\bibinfo{year}{1989}).
\newblock


\bibitem[\protect\citeauthoryear{Xiong, Wang, Zhang, and Li}{Xiong
  et~al\mbox{.}}{2016}]%
        {xiong}
\bibfield{author}{\bibinfo{person}{X. Xiong}, \bibinfo{person}{J. Wang},
  \bibinfo{person}{F. Zhang}, {and} \bibinfo{person}{K. Li}.}
  \bibinfo{year}{2016}\natexlab{}.
\newblock \showarticletitle{Combining Deep Reinforcement Learning and Safety
  Based Control for Autonomous Driving}.
\newblock \bibinfo{journal}{\emph{arXiv:1612.00147}} (\bibinfo{year}{2016}).
\newblock


\end{thebibliography}
